\def\year{2022}\relax
\documentclass[letterpaper]{article} 
\usepackage{aaai22}  
\usepackage{times}  
\usepackage{helvet}  
\usepackage{courier}  
\usepackage[hyphens]{url}  
\usepackage{graphicx} 
\usepackage{natbib}  
\usepackage{caption} 
\DeclareCaptionStyle{ruled}{labelfont=normalfont,labelsep=colon,strut=off} 
\usepackage{algorithm}
\usepackage{algorithmic}

\usepackage{epsfig}
\usepackage{graphicx}
\usepackage{amsmath}
\usepackage{amssymb}

\usepackage{footnote}
\makesavenoteenv{tabular}
\makesavenoteenv{table}
\makesavenoteenv{table*}

\usepackage{newfloat}
\usepackage{listings}
\floatstyle{ruled}
\newfloat{listing}{tb}{lst}{}
\floatname{listing}{Listing}
\title{Sparse Cross-scale Attention Network for Efficient LiDAR Panoptic Segmentation}
\author{
Shuangjie Xu\textsuperscript{\rm {1$\ddagger$}},
Rui Wan\textsuperscript{\rm 2}, 
Maosheng Ye\textsuperscript{\rm 1},
Xiaoyi Zou\textsuperscript{\rm 2},
Tongyi Cao\textsuperscript{\rm 2}
}
\affiliations{
{\tt\small \textsuperscript{\rm 1}The Hong Kong University of Science and Technology \quad \textsuperscript{\rm 2}DEEPROUTE.AI\\
\{sxubj, myeag\}@connect.ust.hk, \{ruiwan, xiaoyizou, tongyicao\}@deeproute.ai}
}

\usepackage{bibentry}

\begin{document}

\maketitle
\noindent\let\thefootnote\relax\footnotetext{\textsuperscript{$\ddagger$}The work was done during an internship at DEEPROUTE.AI.}

\begin{abstract}
Two major challenges of 3D LiDAR Panoptic Segmentation (PS) are that point clouds of an object are surface-aggregated and thus hard to model the long-range dependency especially for large instances, and that objects are too close to separate each other. Recent literature addresses these problems by time-consuming grouping processes such as dual-clustering, mean-shift offsets, etc., or by bird-eye-view (BEV) dense centroid representation that downplays geometry. However, the long-range geometry relationship has not been sufficiently modeled by local feature learning from the above methods. To this end, we present \textit{SCAN}, a novel sparse cross-scale attention network to first align multi-scale sparse features with global voxel-encoded attention to capture the long-range relationship of instance context, which can boost the regression accuracy of the over-segmented large objects. For the surface-aggregated points, \textit{SCAN} adopts a novel sparse class-agnostic representation of instance centroids, which can not only maintain the sparsity of aligned features to solve the under-segmentation on small objects, but also reduce the computation amount of the network through sparse convolution. Our method outperforms previous methods by a large margin in the SemanticKITTI dataset for the challenging 3D PS task, achieving 1\textit{st} place with a real-time inference speed. 
\end{abstract}

\section{Introduction}

\begin{figure}[t]
\vspace{-10px}
    \centering
    \includegraphics[width=8.5cm]{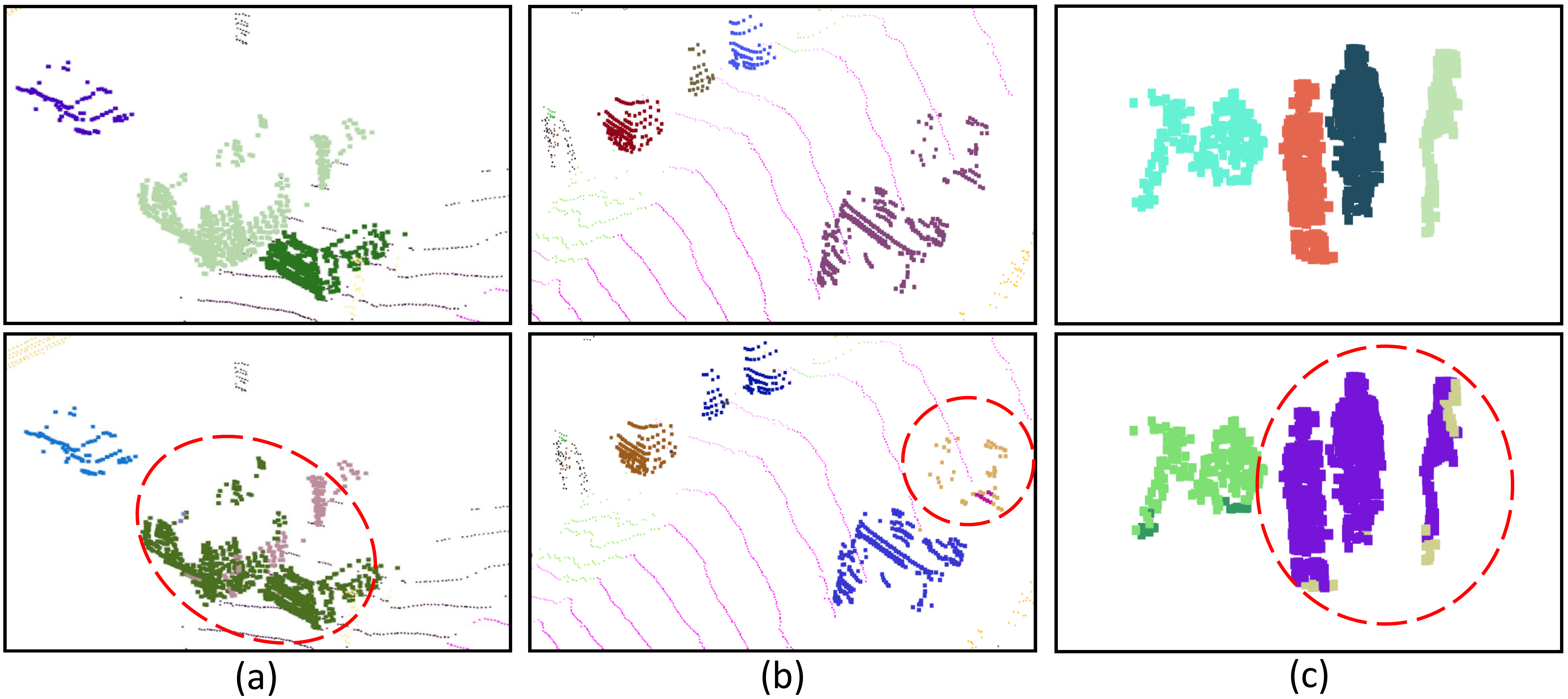}
    \caption{Challenging cases. (a) Two close cars are under-segmented. (b) Surface-aggregated points of a truck are over-segmented. (c) Three pedestrians are merged to one.}
    \label{fig:intro}
    \vspace{-7px}
\end{figure}

3D scene understanding using point clouds has been an essential and challenging task for many robotics applications, including autonomous driving systems. One of the key tasks in 3D scene understanding is 3D Panoptic segmentation (PS), with two sub-tasks, semantic and instance segmentation. The semantic segmentation task aims to attach semantic information at the point level, while the instance segmentation task intends to identify individual countable objects.
Point clouds have sparse, unordered, and irregular-sampled natures and aggregate only on the surface of objects. Such natures pose two main challenges for 3D PS: (1) The surface-aggregated points are far from their object centroids, leading to false segmentation of big objects (Fig.~\ref{fig:intro} a, b); (2) Closely-distributed small objects are falsely merged due to similarity in both physical metric space and feature space (Fig.~\ref{fig:intro} c). Therefore, 
how to correctly group points that belong to individual objects through efficient and effective feature learning becomes a crucial problem.  

\begin{figure}[t]
\vspace{-15px}
    \centering
    \includegraphics[width=8cm]{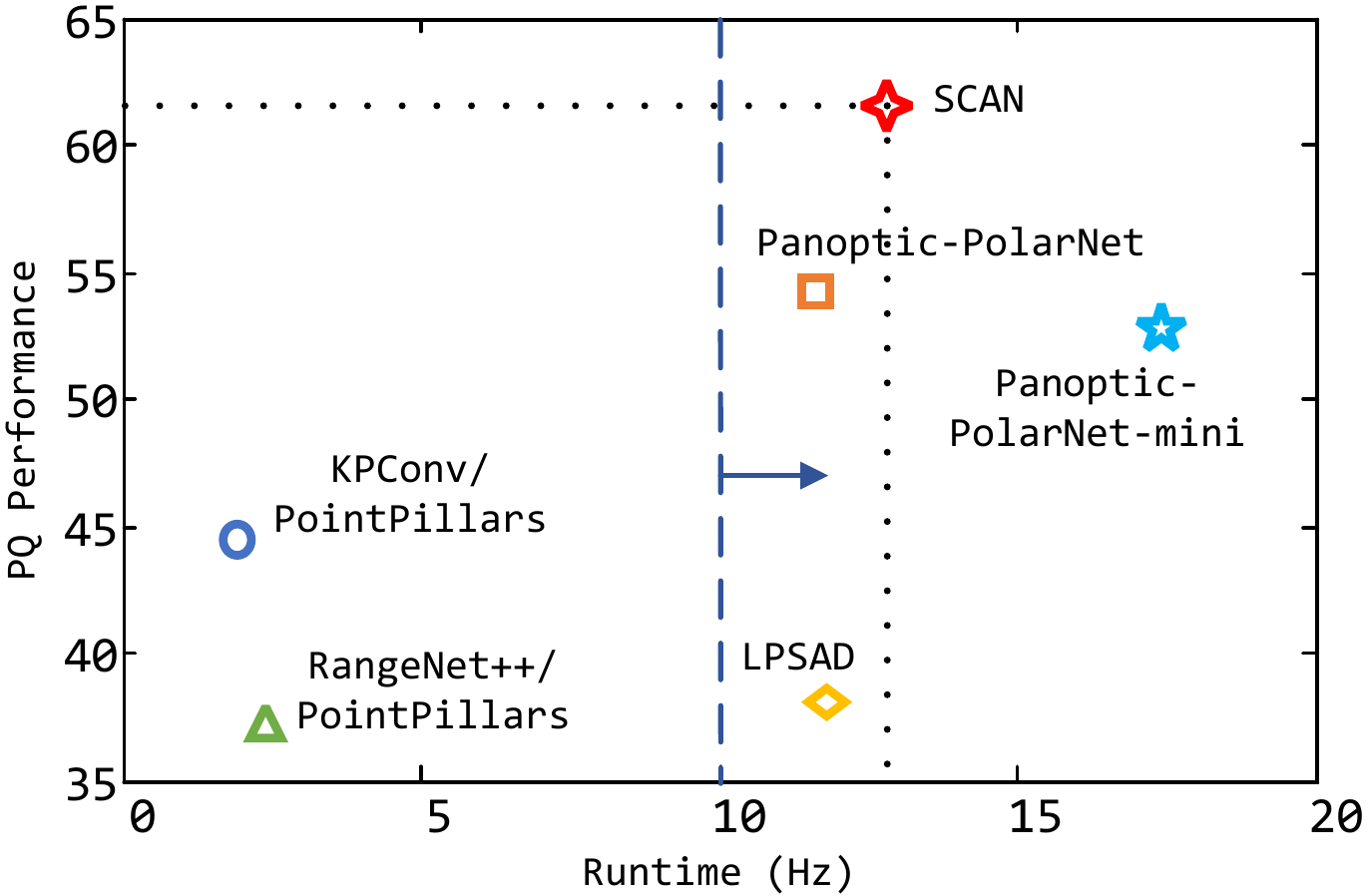}
    \caption{PQ performance vs. inference speed (Hz) in SemanticKITTI panoptic task. The blue vertical line represents the boundary of real-time runtime. The proposed SCAN achieves the state-of-the-art performance.}
    \label{fig:perf}
\vspace{-5px}
\end{figure}

To group the surface-aggregated point clouds, many recent literature~\cite{yang2019learning,liu2020learning,engelmann20203d} adopts the two-stage framework to first propose bounding boxes and then segment instances. To kick over the traces of bounding boxes, most approaches~\cite{wang2019associatively,lahoud20193d} adopt clustering algorithms. VoteNet~\cite{qi2019deep} employs voting to offer broader coverage of ``good" seed points. PointGroup~\cite{jiang2020pointgroup} proposes a dual-clustering method to refine inaccurate offset predictions around object boundaries. DS-Net~\cite{hong2020lidar} utilizes Mean Shift with kernel functions of learned bandwidths to cluster shifted points. 
Such a clustering process requires high time consumption. More importantly, these methods use the center offset as the only guide to instance clustering, making it hard to capture fine-grained long-range geometry relationships and thus resulting in potential over-segmentation of large objects.

Due to the increasing demand for real-time deployment in the industry and academia, plenty of research works focus on efficient LiDAR 3D PS with range or BEV representations.
LPSAD~\cite{milioto2020lidar} and EfficientLPS~\cite{sirohi2021efficientlps} transfer 3D point clouds into range images and achieve high inference speed through 2D convolutional networks.
Panoptic-PolarNet~\cite{zhou2021panoptic} uses polar BEV grids to circumvent the issue of occlusion among instances. 
The above two methods compress the data from 3D to dense 2D representations, leading to the loss of original 3D correlation and the waste of most computation in the empty grids.
Besides, convolution layers applied on dense representations spread information to the invalid grids, making false centroid predictions on occlusion instances and causing under-segmented issues.

Recent works have made great achievements with the sparse voxel architecure~\cite{cheng20212, tang2020searching}, showing the significance of sparsity to point clouds. The widely-used sparse convolutions~\cite{graham20183d} focus on the valid voxels to reduce computation and to avoid dilating the sparse features to invalid voxels. However, the internal association among voxels is difficult to capture with sparse convolutions, especially when the kernel size fails to cover voxels with long-range intervals. Such a ``long-range'' effect is not obvious in the semantic task while critical to the instance task. This can be mitigated by projecting sparse features to the 2D dense representation with 2D convolutions diffusing information, which is, however, neither efficient nor effective. 
To this end, we propose our efficient sparse cross-scale attention network (\textit{SCAN}) that directly models the long-range relationship by a \textit{cross-scale global attention} module. This bottom-up attention mechanism aggregates low-scale, geometrically strong features with high-scale, geometrically weak features, tackling the surface-aggregated problem with multi-scale internal voxel dependency. Besides, we propose \textit{multi-scale sparse supervision} to provide fine-grained features for the attention module.

We also explore the under-segmented issue for occlusion among instances. Instead of timing-consuming clustering-based approach, we follow recent literature~\cite{zhou2021panoptic, cheng2020panoptic} to use the BEV centroid representation. Compared with the existing 3D sparse and 2D dense centroid representations, we propose the \textit{BEV sparse distribution} as our instance centroid prediction for the first time, which keeps the sparsity and the aligned long-range geometry relationships while ensures efficiency at the same time.
Experimental results validate that our \textit{SCAN} is effective and efficient: our method achieves the best performance among the published papers with a real-time speed.
Our main contributions are summarized as follows:
\begin{itemize}
    \item We present \textit{SCAN}, a novel sparse cross-scale attention network to first address the surface-aggregated problem by our \textit{cross-scale global attention} module that directly models long-range dependencies of sparse voxels. 
    \item We propose \textit{multi-scale sparse supervision} to obtain fine-grained features for the cross-scale attention.
    \item We propose \textit{BEV sparse distribution} for centroid prediction for the first time, which boosts performance for instance occlusion and ensures time efficiency.
    \item Our method achieves the best performance among the published papers in the SemanticKITTI dataset for the challenging 3D PS task with a real-time inference speed.
\end{itemize}

\section{Related Work}
This section briefly summarizes recent research related to our work, including 3D semantic segmentation, 3D panoptic segmentation, an attention mechanism.

\noindent \textbf{3D Semantic Segmentation.} Point-based methods~\cite{qi2017pointnet,qi2017pointnet++, li2018pointcnn,hu2020randla,thomas2019kpconv} take raw point clouds as input. They usually sample key points and rely on set abstraction to aggregate local features. However, sampling leads to information loss, and set abstraction 
is computationally costly.
To save computational cost, regular representations, including 3d voxels and 2d grids, polar and cylinder grids, and range images~\cite{zhou2018voxelnet, zhang2020polarnet, zhu2020cylindrical, milioto2019rangenet++, xu2020squeezesegv3} are used to organize sparse points. 
Recently, hybrid methods~\cite{tang2020searching, xu2021rpvnet, ye2021drinet} that combine multiple representations are proposed to integrate the advantages of both fine-grained point-wise features and effective feature aggregation of regular representations. Sparse convolution~\cite{graham2015sparse, graham20183d} is also widely used to restrict convolution output only in the active regions, accelerating the volumetric convolution and enabling larger model size. 

\noindent \textbf{3D Panoptic Segmentation.} 
Current 3D panoptic segmentation methods usually consist of a semantic branch 
and an instance branch.
LPSAD~\cite{milioto2020lidar} obtains instances by clustering shifted points in range images. DS-Net~\cite{hong2021lidar} adaptively and iteratively clusters the learned point-wise centers. 
Based on KPConv~\cite{thomas2019kpconv}, PanosterK~\cite{gasperini2020panoster} introduces the impurity loss and fragmentation loss to train semantic and instance branches jointly, and outputs instance ids straightway from the network. 4D Panoptic~\cite{aygun20214d} proposes a density-based clustering as the initialization and refines it based on the temporal-spatial consistency. Panoptic-PolarNet~\cite{zhou2021panoptic} proposes the 2d dense center heatmap and instance offsets heads for proposal-free instance regression. Major voting is widely adopted as the post-processing to unify the final predictions.

\begin{figure*}[t]
\vspace{-20px}
    \centering
    \includegraphics[width=1.0\textwidth]{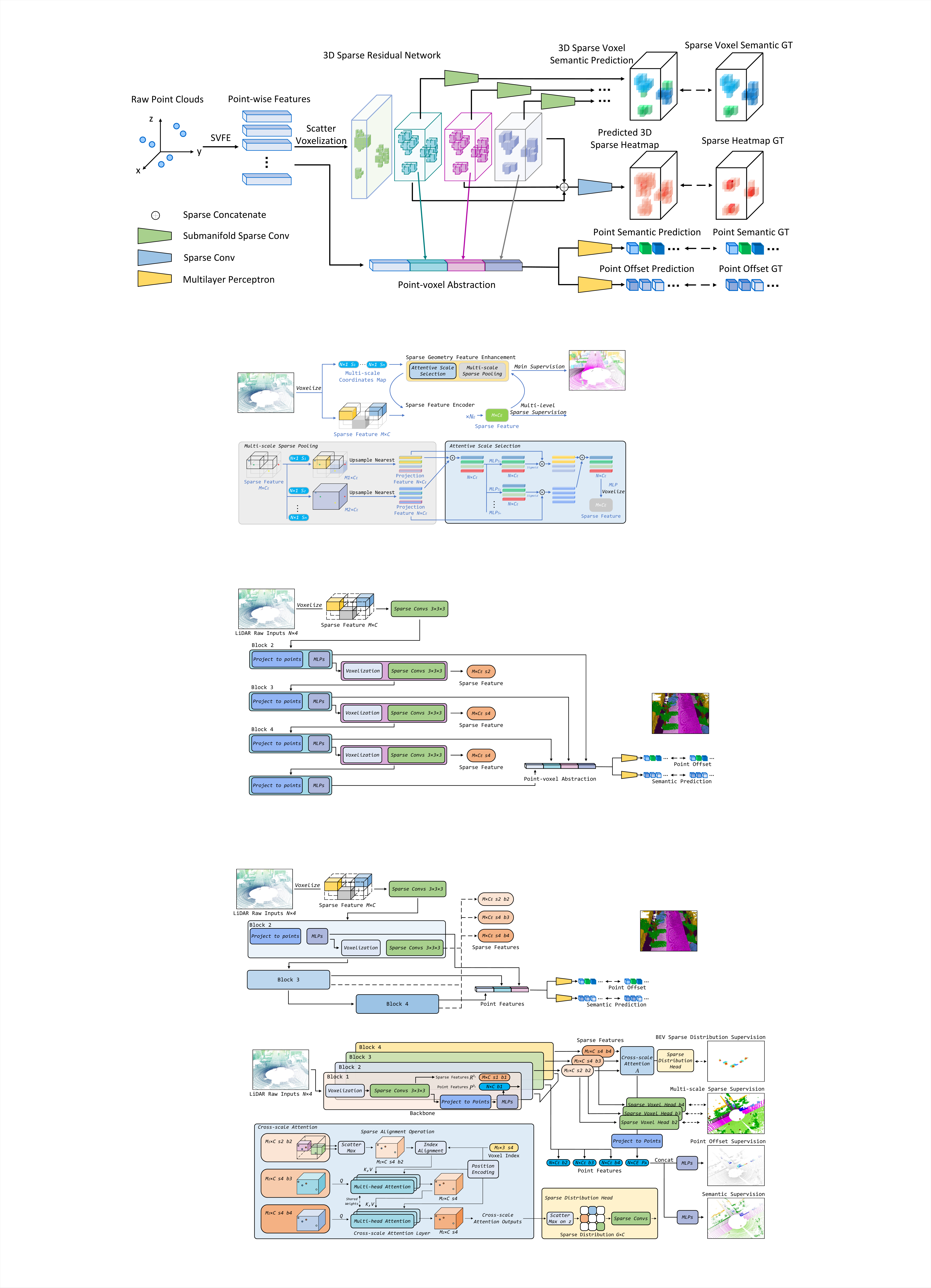}
    \caption{The overall network. Each block of network backbone encodes a sparse voxel feature and a point-wise feature. The point-wise feature is propagated into the next block. Voxel features from the last three blocks are aggregated by the proposed \textit{cross-scale attention} module to acquire the \textit{BEV sparse centroid distribution}. Besides, we apply \textit{multi-scale sparse supervision} on voxel features directly for superior feature learning. Point features from the last three blocks and attention features are concatenated for point-wise offsets from centroids and semantic predictions.
    }
    \label{fig:network}
    \vspace{-5px}
\end{figure*}

\noindent \textbf{Attention Mechanism.}
Attention is defined as the weighted sum of features at multiple positions. SENet ~\cite{hu2018squeeze}, CBAM~\cite{woo2018cbam}, and non-local operation~\cite{wang2018non} have been proposed to exploit channel-wise and spatial attention to adaptively refine features and capture long-range dependencies, which are effective plug-ins for various computer vision tasks, including classification, detection, and segmentation. Transformer~\cite{vaswani2017attention} and DETR~\cite{carion2020end} are the pioneers that rely entirely on attention mechanisms to draw global dependencies between inputs and outputs by stacking self-attention and cross-attention modules. Transformer architectures have also been applied to some indoor point cloud tasks, including classification and segmentation~\cite{engel2020point, zhao2020point, pan20213d, guo2021pct}. However, their memory and computation complexity boost at vast key element numbers and thus hinder the model scalability. Therefore, the variants including deformable attention modules~\cite{zhu2020deformable} and linear attention modules~\cite{katharopoulos2020transformers, choromanski2020rethinking} have been proposed to reduce the computation by utilizing deformable convolutions and matrix properties, respectively.

\section{Approach}

The overall network is illustrated in Fig.~\ref{fig:network}.
The raw point clouds are fed to our network backbone that is composed of four blocks. Each block takes a point-wise feature as the input, and outputs a point-wise feature and sparse voxel feature.
The point-wise feature is propagated into the next block.
Voxel features from the last three blocks are aggregated by the proposed \textit{cross-scale attention} module (Sec.~1) to acquire the \textit{BEV sparse centroid distribution} (Sec.~2). Moreover, we apply \textit{multi-scale sparse supervision} (Sec.~2) on voxel features directly for superior feature learning. Addition network details are described in Sec.~3.
Besides, point features from the last three blocks and attention features are concatenated for point-wise offsets from centroids and semantic predictions.

\noindent \textbf{\textit{Backbone Architecture}}. 
The raw LiDAR inputs $P \in {\mathbb{R}^{N \times 4}}$ ($xyz$ and intensity) are first fed into the network backbone that is composed of four blocks. Each block $b_n,n=1,2,3,4$ first voxelizes the input point features $P^{b_n}$ under voxel size $s$ and encodes them to sparse features ${R^{{b_n}}}{\text{ = }}\left\{ {{I^{{b_n}}} \in {\mathbb{R}^{{M} \times 3}},{F^{{b_n}}} \in {\mathbb{R}^{{M} \times {C}}}} \right\}$ by Submanifold Sparse Convolutional (SSC) layers~\cite{graham20183d}, where $I^{b_n}$ is the coordinate indexes of valid voxels in the form of $x' = \left\lfloor {x/s} \right\rfloor,y' = \left\lfloor {y/s} \right\rfloor,z' = \left\lfloor {z/s} \right\rfloor$ with floor operation $\lfloor \rfloor$, and $F^{b_n}$ denotes feature tensors corresponding to $I^{b_n}$. $M$ denotes valid voxel numbers under current voxel size and $C$ is the channel number of the tensor. Then sparse features are projected back to point features that flow to the next block by Multi-layer Perceptions (MLPs). The voxel size of block $b_{1 \to 4}$ is set to $s, 2s, 4s$ and $4s$ respectively, where $s$ denotes the size of voxels in $x,y,z$-axis measured in meters.
Inspired by great progress achieved by~\cite{ye2021drinet, tang2020searching}, we recognize the importance of multiple representation learning that helps extract better context information. Based on DRINet~\cite{ye2021drinet}, we utilize both point-wise features and sparse voxel features. The point-wise features from multiple blocks are fused to generate point-wise offset with its corresponding semantic prediction, and voxel-wise features can provide instance-level prediction and semantic prediction in sparse format with our proposed cross scale attention.

\noindent \textbf{\textit{Task Abstract}}.
The 3D panoptic segmentation task is abstracted into three sub-tasks inspired by~\cite{zhou2021panoptic}: the point-wise semantic predictions, BEV centroid distribution and centroid-related point offsets, which is a preferred pipeline to take advantage of the voxel-wise features for the long-range relationship acquisition. 
In this work, we abstract the pipeline with four heads: 1) BEV Sparse Distribution head for instance heatmap prediction; 2) Multi-scale Sparse head for auxiliary supervision; 3) Point Offset head and 4) Point-wise Semantic head. For the first two heads, we use sparse voxel features from multi-block $R^{b_n}$ with an attention feature $A$, and for the last two heads, we use point-wise features $P^{b_n}, P^A$ projected from corresponding sparse features, as shown in Fig.~\ref{fig:network}.

The \textit{cross-scale attention} module takes $R^{b_n}$ under multiple scales $2s, 4s$ and multiple levels $b_2, b_3, b_4$ as inputs, outputting aligned and fused sparse features $A$ with internal association among voxels. 
Furthermore, the additional sparse voxel semantic predictions from $b_{2 \to 4}$ are supervised only in training to instruct superior features for feeding into the \textit{cross-scale attention}. To keep the internal association, the \textit{BEV sparse centroid distribution} is adopted instead of the dense version in previous literature.

\subsection{Cross-scale Global Attention}
\label{sec:cross_scale_attention}
In this module, we first align the sparse features from different scales by the proposed \textit{sparse alignment operation}. Then the multi-scale features with aligned coordinate encoding are fed into the \textit{cross-scale global attention layer} that scores the relevance between each pair of voxels and captures the global geometry relationships between the source and the target sparse features. Furthermore, we propose the \textit{multi-scale sparse supervision} to enhance the sparse features as the inputs to the cross-scale attention.

\noindent \textbf{\textit{Sparse Alignment Operation}}.
Shown as Fig.~\ref{fig:network}, sparse features from different voxel scales have different voxel coordinates. Therefore, we propose the sparse alignment operation ($\operatorname{SA}$) to align $R^{b_2}$ at scale $2s$ with $R^{b_3}, R^{b_4}$ at scale $4s$:
\begin{gather}
\begin{gathered}
{{{\bar I}^{{b_2}}}} = \left\lfloor {{{{I^{{b_2}}}} \mathord{\left/
 {\vphantom {{{I^{{b_2}}}} {\left( {{{{4s}} \mathord{\left/
 {\vphantom {{{4s}} {{2s}}}} \right.
 \kern-\nulldelimiterspace} {{2s}}}} \right)}}} \right.
 \kern-\nulldelimiterspace} {\left( {{{{4s}} \mathord{\left/
 {\vphantom {{{4s}} {{2s}}}} \right.
 \kern-\nulldelimiterspace} {{2s}}}} \right)}}} \right\rfloor ,{I^a} = \operatorname{Unique} \left( {{{\bar I}^{{b_2}}}} \right) \\
F^a = {{\max}_j}\left( {{F^{{b_2}}}} \right),\forall j \in {I^{{a}}} \\
{{R^a}} = {\operatorname{SA}}\left( {{R_{{b_2}}}} \right) = \left\{ {{I^{a}},{F^a}} \right\}
\end{gathered}
\end{gather}
, where the function $\operatorname{Unique}$ returns the unique coordinates of input $I$ and ${\max}_j$ is over $F^{b_2}$ whose corresponding coordinates in ${\bar I}^{{b_2}}$ are $j$. The coordinate ${I^a}$ of aligned sparse voxel feature ${R^a}$ is obtained by first downscaled from $2s$ to $4s$ and then $\operatorname{Unique}$ is applied to merge duplicated coordinates. The aligned tensor ${F^a}$ is aggregated over same voxel indexes under scale $4s$. 
However, the aggregated feature ${R^a}$ still doesn't align with $R^{b_3}$ voxel by voxel, because the order of valid voxels may differ between ${I^a}$ and $I^{b_3}$. Therefore, the additional operation $\operatorname{Rearrange}$ named \textit{Voxel-wise Rearrangement} is proposed to reorder ${I^a}$ and ${F^a}$ with the queried mask on hashed coordinates:
\begin{gather}
\begin{gathered}
  {H^{{b_3}}} = {\text{Hash}}\left( {{I^{{b_3}}}} \right),{H^a} = {\text{Hash}}\left( {{{I^a}}} \right) \\
  E = \operatorname{HashQuery} \left( {{H^{{b_3}}},{H^a}} \right) \\
  {{\bar R}^{a}} = \left\{ {\operatorname{Index} \left( {{{I^a}},E} \right),\operatorname{Index} \left( {{{F^a}},E} \right)} \right\}
\end{gathered}
\end{gather}
, where the $j$th element $E_j = i$ of index mask $E$ indicates the index $i$ of $I^a$ whose hashed value $H_i^a = H_j^{{b_3}}$. Then we use $E$ to index $I^a$ and $F^a$ by $\operatorname{Index}$ function to obtain the rearranged feature ${\bar R}^{a}$ whose order is the same as $R^{b_3}$.

\noindent \textbf{\textit{Cross-scale Global Attention Layer}}.
We propose \textit{cross-scale global attention layer} to exploit the inherent multi-scale property, which is proved to be crucial to sparse voxel representation~\cite{ye2021drinet,xu2021rpvnet}. Global attention has been applied for 2D computer vision and indoor point cloud tasks~\cite{vaswani2017attention,pan20213d}.
However, the core issue of applying global attention for long-range relationships on large-scale point clouds is that it would look over all valid voxels. When the voxel scale is low, the number of sparse valid voxels is usually on the order of ten or hundred thousands, making the computation and memory unbearable. Hence, we apply the attention layer on the high-level sparse voxel features by aligning them to the same scale $4s$. To further reduce the computation, we adopt the \textit{Generalized Kernelizable Attention} (GKA)~\cite{choromanski2020rethinking} as our implementation. 

Given an input sparse feature $R^{b_{n}}$ from current block $b_{n}$, let the aligned sparse context feature $R^a$ from $R^{b_{n-1}}$ be encoded as key and value, the cross-scale attention feature $A$ is calculated by:
\begin{equation}
\begin{gathered}
  {{\bar R}^{{b_n}}},{{\bar R}^{a}} = \operatorname{PosEnc} \left( {{R^{{b_n}}}} \right),\operatorname{PosEnc} \left( {{R^a}} \right) \\
  A = \operatorname{GKA} \left( {{{\bar R}^{{b_n}}},\operatorname{MLP} \left( {{{\bar R}^{a}}} \right),\operatorname{MLP}'\left( {{R^a}} \right)} \right)
\end{gathered}
\end{equation}
, where the three inputs of $\operatorname{GKA}$ are \textit{Query}, \textit{Key} and \textit{Value} features respectively. To model the voxel position information into \textit{Query} and \textit{Key} features, we employ a \textit{3D position embedding} fuction $\operatorname{PosEnc}$. For dimension $x$ in $I$, we use commonly used embedding functions~\cite{vaswani2017attention}:
$PE_{2i}^x = \sin \left( {{x \mathord{\left/
 {\vphantom {x {{{10000}^{{{2i} \mathord{\left/
 {\vphantom {{2i} {{d_{{\text{model}}}}}}} \right.
 \kern-\nulldelimiterspace} {{d_{{\text{model}}}}}}}}}}} \right.
 \kern-\nulldelimiterspace} {{{10000}^{{{2i} \mathord{\left/
 {\vphantom {{2i} {{d_{{\text{model}}}}}}} \right.
 \kern-\nulldelimiterspace} {{d_{{\text{model}}}}}}}}}}} \right) $ and $
PE_{2i + 1}^x = \cos \left( {{x \mathord{\left/
 {\vphantom {x {{{10000}^{{{2i} \mathord{\left/
 {\vphantom {{2i} {{d_{{\text{model}}}}}}} \right.
 \kern-\nulldelimiterspace} {{d_{{\text{model}}}}}}}}}}} \right.
 \kern-\nulldelimiterspace} {{{10000}^{{{2i} \mathord{\left/
 {\vphantom {{2i} {{d_{{\text{model}}}}}}} \right.
 \kern-\nulldelimiterspace} {{d_{{\text{model}}}}}}}}}}} \right)$
, where ${d_{{\text{model}}}}{\text{ = }}{{{C_A}} \mathord{\left/
 {\vphantom {{{C_A}} 3}} \right.
 \kern-\nulldelimiterspace} 3}$ and $C_A$ denotes the attention embedding length, $i$ denotes the $i$-th position along the feature channel. The same embedding function is applied on $y$ and $z$. The final position embedding $PE \in {\mathbb{R}^{{M_{{s_i}}} \times {C_A}}}$ from input voxel coordinates $I \in {\mathbb{R}^{{M_{{s_i}}} \times 3}}$ is obtained by setting the first $d_{{\text{model}}}$ channels to $PE^x$, the middle $d_{{\text{model}}}$ channels to $PE^y$ and the left channels to $PE^z$. Finally, $PE$ is applied to \textit{Query} and \textit{Key}.
 
Shown as Fig.~\ref{fig:network}, the bottom-up cross-scale attention starts from aggregating $R^{b_2}$ and $R^{b_3}$ to sparse attention feature $A_1$, and then $A_1$ under scale $4s$ is aggregated with $R^{b_4}$ for the attention output $A$. After our attention module, $A$ now contains the internal voxel relationship cross multi-scale, and is used for the following BEV sparse centroid distribution.

\noindent \textbf{\textit{Multi-scale Sparse Supervision}}.
Existing works make hard semantic labels to supervise dense voxels, which not only costs a large memory footprint but also ignores the possible diversity of point labels within the voxels. Therefore, we supervise multi-scale sparse features from block $b_{2 \to 4}$ directly with soft voxelized semantic labels, shown as ``Multi-scale Sparse Supervision'' in Fig.~\ref{fig:network}. 
By making statistics of semantic labels of points in each voxel, the proportion of each category is taken as the semantic label of the voxel, which constructs the sparse voxel labels $S^{b_n}$. After a few SSC layers on $R^{b_n}$, we use the above-mentioned $\operatorname{Rearrange}$ to align the coordinates from sparse voxel prediction to the sparse label $S^{b_n}$. We employ the L1 loss to obtain the loss of the semantic voxel:
\begin{equation}
    L_v = \sum\limits_{n = 2 \to 4} {{L_{\text{L1} }}\left( {\operatorname{Rearrange} \left( {{\text{SSC}}\left( {{R^{{b_n}}}} \right)} \right),{S^{{b_n}}}} \right)}.
\end{equation}
A ``hard" method takes the major vote category within the voxel as the voxel's category. On the contrary, our ``soft" method calculates statistics of the point number for each category, where the target to regress is $N$ class ratio values for each voxel, making it a regression task. Therefore, we chose the L1 loss instead of a classification loss.

\subsection{BEV Sparse Centroid Distribution}
\label{sec:bev_sparse_representation}
Many previous methods model instance segmentation based on points or centroids. Since instances are spatially separable, the discretized centroid representation~\cite{zhou2021panoptic} is highly suitable for LiDAR instances. Therefore, we choose to use the efficient BEV sparse centroid distribution. In this section, we rethink several possible representations and their pros and cons:

\noindent \textbf{\textit{BEV Dense Distribution}}.
As early adopted in the 2D panoptic task, some works apply the dense distribution to the 3D tasks under BEV~\cite{zhou2021panoptic, ge2020afdet}. The discretized BEV centroid distribution removes the $z$-axis degree of freedom (DOF) and can naturally apply 2D convolutions. However, the dense distribution wastes computation on invalid positions that occupy the majority of the BEV map, especially for heavy network heads. Besides, the 2D convolutions diffuse the captured sparse relationship, which is harmful to our network.

\noindent \textbf{\textit{3D Sparse Distribution}}.
The core issues of 3D sparse distribution are twofold: 1) the load of computation/memory is heavy; 2) the $z$-axis DOF makes the task more difficult compared with the BEV distribution. The benefit is that it can make better use of the geometry information.

\noindent \textbf{\textit{BEV Sparse Distribution}}.
According to the above rethinking, we propose to use the BEV sparse distribution to model the instance centroids in point clouds, which can maintain the sparsity and internal relationship in voxel features while keep efficiency by only computing valid BEV positions through SSC. With the attention output $A$, we first set all ${z \to 0}$ of $I^A$ to flatten the $z$-axis. Then we apply $\operatorname{Unique}$ on the new coordinates and obtain the max features over BEV unique voxels by $\max _j$ operator in Equ.~1, which generates the BEV sparse feature ${\bar A}$. With several SSC layers, the final BEV sparse distribution is obtained as $D \in {\mathbb{R}^{w \times h}}$ where $w$ and $h$ denote grid sizes in $x$-axis and $y$-axis.

\subsection{Network Details}
\label{sec:network_detail}

\noindent \textbf{\textit{Supervision}}.
According to our task abstract, we divide 3D panoptic task into three sub-tasks: 1) BEV sparse centroid distribution prediction $D$ supervised with Focal loss~\cite{lin2017focal} $L_d$; 2) point-wise offset prediction $O \in {\mathbb{R}^{N \times 2}}$ that denotes the distances on $x,y$ axes respectively between points and the corresponding centroids; 3) point-wise semantic prediction $S \in {\mathbb{R}^{N \times 1}}$. We supervise point-wise $O$ with L1 loss as $L_o$ and $S$ with the sum of the Lov{\'a}sz loss~\cite{berman2018lovasz} and Focal loss as $L_s$. Besides, we mask out the background points during calculating $L_o$. Furthermore, we supervise the multi-scale sparse semantic prediction with L1 loss as an auxiliary loss $L_v$. The total loss is the sum of the above losses:
\begin{equation}
    L = {L_d} + {L_o} + {L_s} + {L_v}
\end{equation}

\noindent \textbf{\textit{Panoptic Inference}}.
During inference, to further obtain the centroid prediction, we first apply sparse max pooling on $D$ and then keep the voxel coordinates with unchanged features before and after this pooling. We keep $K$ centroids with top confidence scores as the final centroid predictions. 

By the point-wise semantic predictions $S$, we get the \textit{thing} points. By the predicted $K$ centroids and point-wise offsets $O$, we 
shift each \textit{thing} point and then assign each shifted point to its closest centroid to get clustering results. Since $K$ is set to cover the max number of instances, some predicted centroids are assigned with no points, which are removed during inference. To further refine panoptic results, we obtain the semantic label of each centroid by majority voting within the semantic predictions of its associated points, then we relabel the outlier points in each voxel. Besides, the instance IDs of \textit{stuff} points are set to $0$.


\begin{figure*}[t]
\vspace{-20px}
    \centering
    \includegraphics[width=1.0\textwidth]{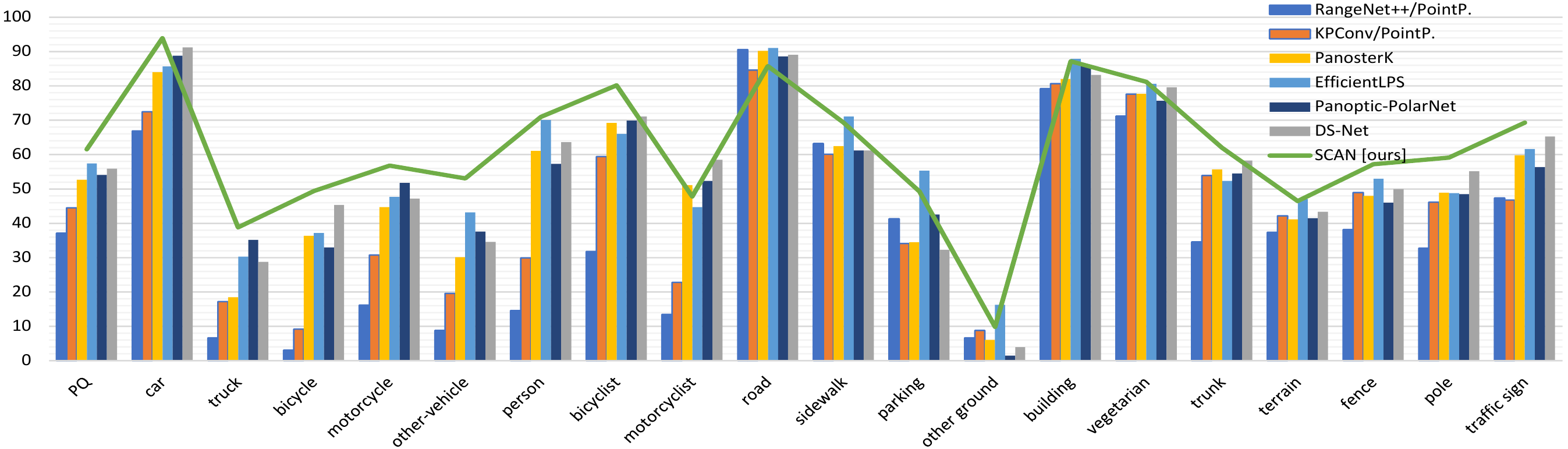}
    \caption{Per-class results of metric PQ on the SemanticKITTI test set.}
    \label{fig:cls_results}
\end{figure*}

\begin{table*}[t]
\small
\centering
\renewcommand{\arraystretch}{1.1}
\begin{tabular}{l|cccc|ccc|ccc|c|c}
Method                                                                           & PQ            & $\text{PQ}^\dagger$ & SQ            & RQ            & $\text{PQ}^{th}$ & $\text{SQ}^{th}$ & $\text{RQ}^{th}$ & $\text{PQ}^{st}$ & $\text{SQ}^{st}$ & $\text{RQ}^{st}$ & mIoU          & FPS         \\ \hline \hline
RangeNet++/PointPillars & 37.1          & 45.9                & 75.9          & 47.0          & 20.2             & 75.2             & 25.2             & 49.3             & 76.5             & 62.8             & 52.4          & 2.4         \\
KPConv/PointPillars          & 44.5          & 52.5                & 80.0          & 54.4          & 32.7             & 81.5             & 38.7             & 53.1             & 79.0    & 65.9             & 58.8          & 1.9         \\
LPSAD   & 38.0          & 47.0                & 76.5          & 48.2          & 25.6             & 76.8             & 31.8             & 47.1             & 76.2             & 60.1             & 50.9          & 11.8        \\
Panoster     & 52.7          & 59.9                & 80.7          & 64.1 & 49.4             & 83.3             & 58.5             & 55.1    & 78.8             & 68.2    & 59.9          &  -         \\
DS-Net        & 55.9          & 62.5                & 82.3          & 66.7 & 55.1             & 87.2             & 62.8             & 56.5    & 78.7             & 69.5    & 61.6          &  -         \\
Panoptic-PolarNet        & 54.1         &  60.7                & 81.4         &  65.0 & 53.3            & 87.2            & 60.6             & 54.8    &  77.2             &  68.1    & 59.5          &  11.6     \\
EfficientLPS        & 57.4         &  63.2                & 83.0          &  68.7 & 53.1            & 87.8             & 60.5             & 60.5    &  79.5             &  \textbf{74.6}    & 61.4          &  -         \\
GP-S3Net        & 60.0         &  \textbf{69.0}               & 82.0          &   \textbf{72.1} & \textbf{65.0}            & 86.6             &  \textbf{74.5}             & 56.4    &  78.7             &  70.4    & \textbf{70.8}          &  -         \\ 
\hline \hline
SCAN    & \textbf{61.5}       & 67.5      & \textbf{84.5}        & \textbf{72.1}       & 61.4            & \textbf{88.1}            & 69.3             & \textbf{61.5}          & \textbf{81.8}             & 74.1     & 67.7     & \textbf{12.8} \\
\end{tabular}
\caption{Comparison of test set results on SemanticKITTI using stuff (st) and thing(th) classes. All results in [\%].}
\label{tab:main_results}
\end{table*}

\begin{table*}[t]
\vspace{-15px}
\small
\centering
\renewcommand{\arraystretch}{1.1}
\begin{tabular}{l|cccc|ccc|ccc|c}
Method            & PQ   & $\text{PQ}^\dagger$ & SQ   & RQ   & $\text{PQ}^{th}$ & $\text{SQ}^{th}$ & $\text{RQ}^{th}$ & $\text{PQ}^{st}$ & $\text{SQ}^{st}$ & $\text{RQ}^{st}$ & mIOU \\ \hline \hline
DS-Net            & 42.5 & 51.0                & 50.3 & 83.6 & 32.5             & 83.1             & 38.3             & 59.2             & 84.4             & 70.3             & 70.7 \\
EfficientLPS      & 59.2 & 62.8                & 82.9 & 70.7 & 51.8             & 80.6             & 62.7             & 71.5             & 84.3             & 84.1             & 69.4 \\
\hline \hline
SCAN(ours)        & \textbf{65.1} & \textbf{68.9}       & \textbf{85.7} & \textbf{75.3} & \textbf{60.6}    & \textbf{85.7}    & \textbf{70.2}    & \textbf{72.5}    & \textbf{85.7}    & \textbf{83.8}    & \textbf{77.4}\\
\end{tabular}
\caption{Comparison of validation set results on Nuscenes using stuff (st) and thing(th) classes. All results in [\%].}
\label{tab:nuscenes_results}
\end{table*}

\begin{figure}[htbp]
    \centering
    \includegraphics[width=8cm]{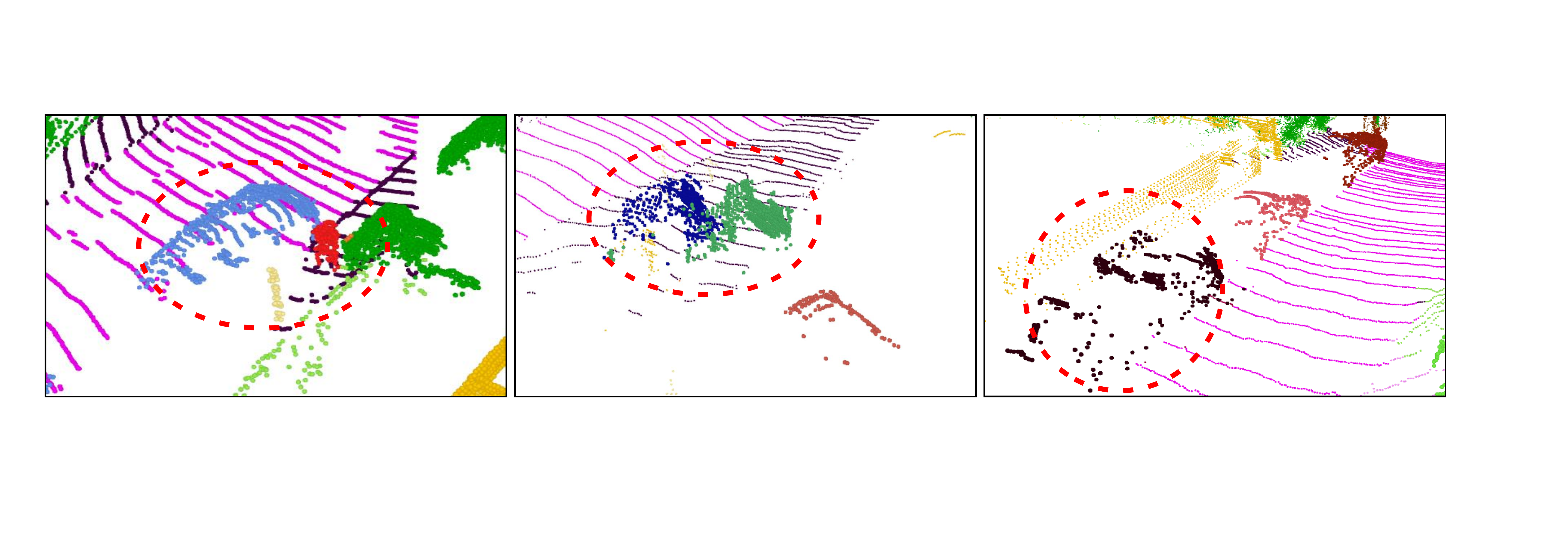}
    \caption{Results of \textit{SCAN}. The first two samples demonstrate its effectiveness for challenging under-segmentation, and the last shows the robustness on over-segmentation.}
    \label{fig:good_sample}
\end{figure}

\section{Experiments}

In this section, we investigate our method's performance on the standard benchmark dataset SemanticKITTI~\cite{behley2020benchmark} and Nuscenes~\cite{nuscenes2019}. We compare our model with state-of-the-art methods and perform an ablation study to demonstrate the advantage of each module in \textit{SCAN}.

\noindent \textbf{SemanticKITTI.} SemanticKITTI~\cite{behley2019semantickitti,behley2020benchmark} is a challenging dataset, proposed to provide full 360-degree point-wise labels for the large-scale LiDAR data of the KITTI Odometry Benchmark~\cite{geiger2012we}. It contains $23201$ scans with 3D semantic and instance annotations for training and $20351$ for testing. The test evaluation is on the official server with 11 \textit{stuff} classes and 8 \textit{thing} classes.

\noindent \textbf{Nuscenes.} The large-scale Nuscenes dataset~\cite{nuscenes2019} has newly released the ~\textit{panoptic segmentation} challenge. The annotations include 10 \textit{thing} classes and 6 \textit{stuff} classes out of total 16 semantic classes. The dataset contains 1000 scenes, including 850 scenes for training and validation and 150 scenes for testing. Since the leaderboard has not opened until this paper is submitted, we only use the training and validation set in the experiment, which have 28130 and 6019 frames, respectively.

\noindent \textbf{Evaluation.}
To assess the semantic segmentation, we rely on the commonly-used mean intersection-over-union (mIoU) metric~\cite{behley2019semantickitti} over all classes.
To measure the quality of point cloud panoptic segmentation, we adopt the standard convention~\cite{behley2020benchmark} with PQ, SQ and RQ.

\begin{figure*}[t]
    \centering
    \includegraphics[width=0.9\textwidth]{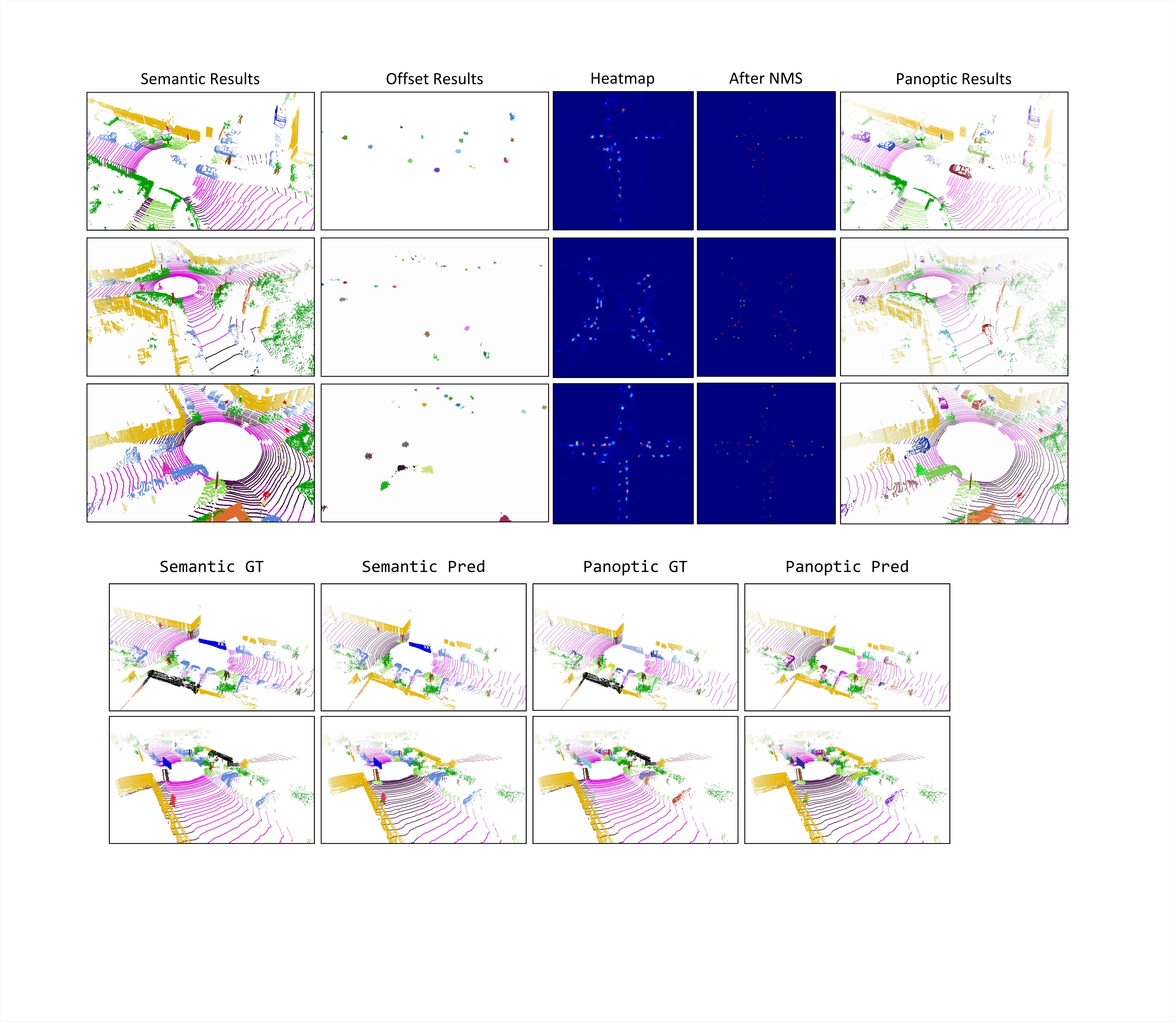}
    \caption{Qualitative results of \textit{SCAN} from the SemanticKITTI validation sets.}
    \label{fig:vis_main}
    \vspace{-5px}
\end{figure*}

\subsection{Implementation Details}

\textbf{Training.}
We fix the voxelization space to be limited in $[[\pm48], [\pm48], [-3,1.5]]$.
We do global rotation along $z$ axis in range of $[-\pi, \pi]$ degrees and flip the points along $x$, $y$, and $x+y$ axes. Each augmentation is applied independently with a probability of 50\%. In addition, we set the default scale $s = \left[ {0.2,0.2,0.1} \right]$ measured in metres, thus the $w=120, h=120$ for the BEV sparse centroid distribution.
The feature channels are set to $C=64$ in the network, and we configure the \textit{GKA} attention by setting the number of heads $8$, attention depth $2$, channels of each head $16$ and disabling the causality inference.
We implement the network in PyTorch~\cite{paszke2017automatic} and train the \textit{SCAN} model on 8 NVIDIA 3090 GPUs for 40 epochs with Adam~\cite{kingma2014adam} and 1-Cycle Schedule~\cite{smith2017cyclical}. We set the batch size per GPU as 4 and the initial learning rate 0.003. The learning rate first raises tenfold before the 16-th epoch and then decays. 

\noindent \textbf{Inference.}
During inference, the auxiliary voxel semantic prediction is cut off to save the computation.
The max centroid number $K$ is set to 100 and the centroid score threshold is set to $0.1$.
For points out of the fixed voxelization space, we set both the semantic and instance predictions to $0$.

\subsection{Experiment Results}

\textbf{Comparison to the State-of-the-arts.}
Tab.~\ref{tab:main_results} shows the quantitative comparison on the SemanticKITTI test set submitted to the official test server. We find that \textit{SCAN} outperforms GP-S3Net~\cite{razani2021gp} by 1.5\% while ensures real-time inference speed, setting a new state-of-the-art performance for the LiDAR panoptic segmentation.
We present our Nuscenes validation results in Table~\ref{tab:nuscenes_results}, and compare results with models that use the same settings of the dataset.

\noindent \textbf{Inference Speed.}
Tab.~\ref{tab:main_results} also shows the frames per second (FPS) for each approach.
The reported speed for our approach is the full time including all pre- and post- processing.
We investigate that our model can achieve a real-time inference speed for autonomous driving.
Compared with other models that reported speed (containing preprocessing), our approach maintains a faster speed while improves accuracy.

\noindent \textbf{Improvement.}
Many previous works are troubled by over- or under- segmentation problems.
However, as shown in Fig.~\ref{fig:cls_results}, our method handles these challenges well.
In the case of large objects like \textit{truck}s which are often over-segmented, our method outperforms other approaches by a large margin.
Also, we investigate that our method is robust enough to deal with the cases of under-segmentation and small objects, especially for \textit{person}, \textit{other-vehicle} and \textit{motorcyclist} classes. Such improvements demonstrate the delicate localization and effective object grouping of \textit{SCAN}. 

\noindent \textbf{Qualitative Results.}
The qualitative performance on the challenging scans can be illustrated in Fig.~\ref{fig:cls_results}.
Besides, shown as Fig.~\ref{fig:good_sample}, our approach can handle the challenging samples with over- and under- segmentation.

\subsection{Ablation Study.}

To analyze the effectiveness of different modules in \textit{SCAN}, adequate ablation studies are conducted on the SemanticKITTI valid set.
Starting from a baseline whose network backbone is the same with \textit{SCAN}, we switch on proposed modules respectively, as shown in Tab.~\ref{tab:ablation_1}, which indicates the metrics and inference \textit{Runtime} (batch size is 1) of models with NVIDIA 3090 GPU.
Instead of directly summing multi-scale sparse features, the proposed \textit{cross-scale global attention} \textit{CGA} improves by the largest margin of 2.7\% compared to the baseline in PQ with small computation increment, demonstrating the validity.
The \textit{multi-scale sparse supervision} \textit{MSS} achieves a 0.4\% promotion separately but gives 0.7\% gain together with \textit{CGA}, which indicates the proposed \textit{CGA} can be stimulated by fine-grained features.
Notice that \textit{MSS} brings no influence in speed because that \textit{MSS} will be cut off during inference.
We also conduct experiments on different representations of centroid distribution. The performance of the default used dense version \textit{DD} is 0.2\% higher than the 3D sparse version \textit{3SD}. Our \textit{BEV sparse centroid distribution} \textit{SD} achieves the best performance with an even smaller computation.

\begin{table}[t]
\small
    \centering
    \begin{tabular}{cc|ccc|c|c|c}
MSS  &  CGA &  DD & 3SD & SD & PQ  & mIoU  & Runtime \\ \hline
 &  & \checkmark &  &  & 53.3 & 65.8 & 66ms\\
\checkmark &  & \checkmark &  &  & 53.7 & 66.1 & 66ms \\
 & \checkmark & \checkmark  &   &  & 56.0 & 67.9 & 81ms \\ 
\checkmark & \checkmark &   \checkmark  & &  & 56.7 & 68.5 & 81ms \\  \hline
\checkmark & \checkmark &    & \checkmark &  & 56.5 & 68.4 & 85ms \\
\checkmark & \checkmark &    &  & \checkmark & \textbf{57.2} & \textbf{68.9} & 78ms \\

\end{tabular}
\caption{Ablation study of \textit{SCAN} on the validation split of SemanticKITTI. \textit{CGA}, \textit{MSS}, \textit{SD}, \textit{DD}, \textit{3SD} represent for \textit{cross-scale global attention} (\textit{CGA}), \textit{multi-scale sparse supervision} (\textit{MSS}), the \textit{BEV sparse centroid distribution} (\textit{SD}) with its dense version (\textit{DD}) and 3d sparse version (\textit{3SD}).}
\label{tab:ablation_1}
\end{table}

We also conduct experiments on the \textit{cross-scale global attention} module, as shown in Tab.~\ref{tab:ablation_CGA} to investigate the module that brings the most gain. We first adopt various input combinations of sparse features from different blocks $b_{1\to4}$. Results indicate that attention on $b_{2\to4}$ outperforms other combinations. Only taking $b_4$ as input makes the module tune into the self-attention. The biggest improvement occurs between $b_3,b_4$ cross-attention and $b_4$ self-attention, where the former provides a foundation for the acquisition of the internal voxel relationship across sparse features. Including the $b_1$ feature degrades performance, for which the reason may be the insufficient learning of sparse features in the first block. Moreover, we try to remove the 3D position encoding \textit{PE} and the performance has a sharp drop by 1.2\%, demonstrating the importance of 3D voxel coordinates towards global attention learning. Forcing the same attention layer \textit{SW} to learn the long-range relationship gives another 0.6\% improvement. By sharing weights among attention layers, our \textit{cross-scale global attention} module concentrates on the learning of cross-scale attention patterns. 

\begin{table}[t]
\small
    \centering
    \begin{tabular}{cccc|cc|cc}
$b_1$ & $b_2$ & $b_3$ & $b_4$ & PE & SW & PQ & mIoU\\ \hline
 & & & \checkmark & \checkmark & & 54.5 & 67.4 \\
 & & \checkmark & \checkmark & \checkmark & & 55.9 & 68.1 \\
 & \checkmark & \checkmark & \checkmark & \checkmark & & 56.8 & 68.7 \\
\checkmark & \checkmark & \checkmark & \checkmark & \checkmark & & 56.6 & 68.6 \\ \hline
 & \checkmark & \checkmark & \checkmark & & & 55.6 & 68.1 \\
 & \checkmark & \checkmark & \checkmark & \checkmark & \checkmark & \textbf{57.2} & \textbf{68.9} \\

\end{tabular}
\caption{Ablation study of the \textit{cross-scale global attention} module. $b_{1\to4}$ stands features from which block are as inputs to the module. \textit{PE} represents the 3D position encoding and \textit{SW} denotes sharing weight for attention layers.}
\label{tab:ablation_CGA}
\end{table}

\section{Conclusion}
We present efficient \textit{SCAN}, a novel sparse cross-scale attention network to first address the surface-aggregated problem in the 3D panoptic segmentation task by modeling the long-range dependency among sparse voxel representation. The proposed \textit{cross-scale attention} module introduces the attention mechanism to align and fuse multi-level and multi-scale sparse features in global instead of only stacking sparse convolution layers for local context information. Moreover, the \textit{multi-scale sparse voxel supervision} is proposed to obtain fine-grained features for the \textit{cross-scale attention}. In addition, we rethink centroid distributions and finally choose the \textit{BEV sparse distribution} for better performance with lower computation and memory footprint. Our method achieves the state-of-the-art among published work and $1^{st}$ in the SemanticKITTI challenge with a real-time runtime speed.

{\small
\bibliography{egbib}

\begin{thebibliography}{54}
\providecommand{\natexlab}[1]{#1}

\bibitem[{Aygun et~al.(2021)Aygun, Osep, Weber, Maximov, Stachniss, Behley, and
  Leal-Taix{\'e}}]{aygun20214d}
Aygun, M.; Osep, A.; Weber, M.; Maximov, M.; Stachniss, C.; Behley, J.; and
  Leal-Taix{\'e}, L. 2021.
\newblock 4D Panoptic LiDAR Segmentation.
\newblock In \emph{IEEE Conf. Comput. Vis. Pattern Recog.}, 5527--5537.

\bibitem[{Behley et~al.(2019)Behley, Garbade, Milioto, Quenzel, Behnke,
  Stachniss, and Gall}]{behley2019semantickitti}
Behley, J.; Garbade, M.; Milioto, A.; Quenzel, J.; Behnke, S.; Stachniss, C.;
  and Gall, J. 2019.
\newblock SemanticKITTI: A dataset for semantic scene understanding of lidar
  sequences.
\newblock In \emph{Int. Conf. Comput. Vis.}, 9297--9307.

\bibitem[{Behley, Milioto, and Stachniss(2020)}]{behley2020benchmark}
Behley, J.; Milioto, A.; and Stachniss, C. 2020.
\newblock A Benchmark for LiDAR-based Panoptic Segmentation based on KITTI.
\newblock \emph{arXiv preprint arXiv:2003.02371}.

\bibitem[{Berman, Rannen~Triki, and Blaschko(2018)}]{berman2018lovasz}
Berman, M.; Rannen~Triki, A.; and Blaschko, M.~B. 2018.
\newblock The lov{\'a}sz-softmax loss: A tractable surrogate for the
  optimization of the intersection-over-union measure in neural networks.
\newblock In \emph{IEEE Conf. Comput. Vis. Pattern Recog.}, 4413--4421.

\bibitem[{Caesar et~al.(2019)Caesar, Bankiti, Lang, Vora, Liong, Xu, Krishnan,
  Pan, Baldan, and Beijbom}]{nuscenes2019}
Caesar, H.; Bankiti, V.; Lang, A.~H.; Vora, S.; Liong, V.~E.; Xu, Q.; Krishnan,
  A.; Pan, Y.; Baldan, G.; and Beijbom, O. 2019.
\newblock nuScenes: A multimodal dataset for autonomous driving.
\newblock \emph{arXiv preprint arXiv:1903.11027}.

\bibitem[{Carion et~al.(2020)Carion, Massa, Synnaeve, Usunier, Kirillov, and
  Zagoruyko}]{carion2020end}
Carion, N.; Massa, F.; Synnaeve, G.; Usunier, N.; Kirillov, A.; and Zagoruyko,
  S. 2020.
\newblock End-to-end object detection with transformers.
\newblock In \emph{Eur. Conf. Comput. Vis.}, 213--229. Springer.

\bibitem[{Cheng et~al.(2020)Cheng, Collins, Zhu, Liu, Huang, Adam, and
  Chen}]{cheng2020panoptic}
Cheng, B.; Collins, M.~D.; Zhu, Y.; Liu, T.; Huang, T.~S.; Adam, H.; and Chen,
  L.-C. 2020.
\newblock Panoptic-deeplab: A simple, strong, and fast baseline for bottom-up
  panoptic segmentation.
\newblock In \emph{IEEE Conf. Comput. Vis. Pattern Recog.}, 12475--12485.

\bibitem[{Cheng et~al.(2021)Cheng, Razani, Taghavi, Li, and Liu}]{cheng20212}
Cheng, R.; Razani, R.; Taghavi, E.; Li, E.; and Liu, B. 2021.
\newblock 2-S3Net: Attentive feature fusion with adaptive feature selection for
  sparse semantic segmentation network.
\newblock In \emph{IEEE Conf. Comput. Vis. Pattern Recog.}, 12547--12556.

\bibitem[{Choromanski et~al.(2020)Choromanski, Likhosherstov, Dohan, Song,
  Gane, Sarlos, Hawkins, Davis, Mohiuddin, Kaiser
  et~al.}]{choromanski2020rethinking}
Choromanski, K.; Likhosherstov, V.; Dohan, D.; Song, X.; Gane, A.; Sarlos, T.;
  Hawkins, P.; Davis, J.; Mohiuddin, A.; Kaiser, L.; et~al. 2020.
\newblock Rethinking attention with performers.
\newblock \emph{arXiv preprint arXiv:2009.14794}.

\bibitem[{Engel, Belagiannis, and Dietmayer(2020)}]{engel2020point}
Engel, N.; Belagiannis, V.; and Dietmayer, K. 2020.
\newblock Point transformer.
\newblock \emph{arXiv preprint arXiv:2011.00931}.

\bibitem[{Engelmann et~al.(2020)Engelmann, Bokeloh, Fathi, Leibe, and
  Nie{\ss}ner}]{engelmann20203d}
Engelmann, F.; Bokeloh, M.; Fathi, A.; Leibe, B.; and Nie{\ss}ner, M. 2020.
\newblock 3D-MPA: Multi-Proposal Aggregation for 3D Semantic Instance
  Segmentation.
\newblock In \emph{IEEE Conf. Comput. Vis. Pattern Recog.}, 9031--9040.

\bibitem[{Gasperini et~al.(2021)Gasperini, Mahani, Marcos-Ramiro, Navab, and
  Tombari}]{gasperini2020panoster}
Gasperini, S.; Mahani, M.-A.~N.; Marcos-Ramiro, A.; Navab, N.; and Tombari, F.
  2021.
\newblock Panoster: End-to-end Panoptic Segmentation of LiDAR Point Clouds.
\newblock \emph{IEEE Robotics and Automation Letters}.

\bibitem[{Ge et~al.(2020)Ge, Ding, Hu, Wang, Chen, Huang, and Li}]{ge2020afdet}
Ge, R.; Ding, Z.; Hu, Y.; Wang, Y.; Chen, S.; Huang, L.; and Li, Y. 2020.
\newblock Afdet: Anchor free one stage 3d object detection.
\newblock \emph{arXiv preprint arXiv:2006.12671}.

\bibitem[{Geiger, Lenz, and Urtasun(2012)}]{geiger2012we}
Geiger, A.; Lenz, P.; and Urtasun, R. 2012.
\newblock Are we ready for autonomous driving? the kitti vision benchmark
  suite.
\newblock In \emph{IEEE Conf. Comput. Vis. Pattern Recog.}, 3354--3361. IEEE.

\bibitem[{Graham(2015)}]{graham2015sparse}
Graham, B. 2015.
\newblock Sparse 3D convolutional neural networks.
\newblock \emph{Brit. Mach. Vis. Conf.}

\bibitem[{Graham, Engelcke, and Van Der~Maaten(2018)}]{graham20183d}
Graham, B.; Engelcke, M.; and Van Der~Maaten, L. 2018.
\newblock 3d semantic segmentation with submanifold sparse convolutional
  networks.
\newblock In \emph{IEEE Conf. Comput. Vis. Pattern Recog.}, 9224--9232.

\bibitem[{Guo et~al.(2021)Guo, Cai, Liu, Mu, Martin, and Hu}]{guo2021pct}
Guo, M.-H.; Cai, J.-X.; Liu, Z.-N.; Mu, T.-J.; Martin, R.~R.; and Hu, S.-M.
  2021.
\newblock PCT: Point cloud transformer.
\newblock \emph{Computational Visual Media}, 7(2): 187--199.

\bibitem[{Hong et~al.(2020)Hong, Zhou, Zhu, Li, and Liu}]{hong2020lidar}
Hong, F.; Zhou, H.; Zhu, X.; Li, H.; and Liu, Z. 2020.
\newblock LiDAR-based Panoptic Segmentation via Dynamic Shifting Network.
\newblock \emph{arXiv preprint arXiv:2011.11964}.

\bibitem[{Hong et~al.(2021)Hong, Zhou, Zhu, Li, and Liu}]{hong2021lidar}
Hong, F.; Zhou, H.; Zhu, X.; Li, H.; and Liu, Z. 2021.
\newblock Lidar-based panoptic segmentation via dynamic shifting network.
\newblock In \emph{IEEE Conf. Comput. Vis. Pattern Recog.}, 13090--13099.

\bibitem[{Hu, Shen, and Sun(2018)}]{hu2018squeeze}
Hu, J.; Shen, L.; and Sun, G. 2018.
\newblock Squeeze-and-excitation networks.
\newblock In \emph{IEEE Conf. Comput. Vis. Pattern Recog.}, 7132--7141.

\bibitem[{Hu et~al.(2020)Hu, Yang, Xie, Rosa, Guo, Wang, Trigoni, and
  Markham}]{hu2020randla}
Hu, Q.; Yang, B.; Xie, L.; Rosa, S.; Guo, Y.; Wang, Z.; Trigoni, N.; and
  Markham, A. 2020.
\newblock RandLA-Net: Efficient semantic segmentation of large-scale point
  clouds.
\newblock In \emph{IEEE Conf. Comput. Vis. Pattern Recog.}, 11108--11117.

\bibitem[{Jiang et~al.(2020)Jiang, Zhao, Shi, Liu, Fu, and
  Jia}]{jiang2020pointgroup}
Jiang, L.; Zhao, H.; Shi, S.; Liu, S.; Fu, C.-W.; and Jia, J. 2020.
\newblock PointGroup: Dual-Set Point Grouping for 3D Instance Segmentation.
\newblock In \emph{IEEE Conf. Comput. Vis. Pattern Recog.}, 4867--4876.

\bibitem[{Katharopoulos et~al.(2020)Katharopoulos, Vyas, Pappas, and
  Fleuret}]{katharopoulos2020transformers}
Katharopoulos, A.; Vyas, A.; Pappas, N.; and Fleuret, F. 2020.
\newblock Transformers are rnns: Fast autoregressive transformers with linear
  attention.
\newblock In \emph{ICML}, 5156--5165. PMLR.

\bibitem[{Kingma and Ba(2014)}]{kingma2014adam}
Kingma, D.~P.; and Ba, J. 2014.
\newblock Adam: A method for stochastic optimization.
\newblock \emph{arXiv preprint arXiv:1412.6980}.

\bibitem[{Lahoud et~al.(2019)Lahoud, Ghanem, Pollefeys, and
  Oswald}]{lahoud20193d}
Lahoud, J.; Ghanem, B.; Pollefeys, M.; and Oswald, M.~R. 2019.
\newblock 3d instance segmentation via multi-task metric learning.
\newblock In \emph{Int. Conf. Comput. Vis.}, 9256--9266.

\bibitem[{Li et~al.(2018)Li, Bu, Sun, Wu, Di, and Chen}]{li2018pointcnn}
Li, Y.; Bu, R.; Sun, M.; Wu, W.; Di, X.; and Chen, B. 2018.
\newblock Pointcnn: Convolution on x-transformed points.
\newblock In \emph{Adv. Neural Inform. Process. Syst.}, 820--830.

\bibitem[{Lin et~al.(2017)Lin, Goyal, Girshick, He, and
  Doll{\'a}r}]{lin2017focal}
Lin, T.-Y.; Goyal, P.; Girshick, R.; He, K.; and Doll{\'a}r, P. 2017.
\newblock Focal loss for dense object detection.
\newblock In \emph{Int. Conf. Comput. Vis.}, 2980--2988.

\bibitem[{Liu et~al.(2020)Liu, Yu, Wu, Chen, and Liu}]{liu2020learning}
Liu, S.-H.; Yu, S.-Y.; Wu, S.-C.; Chen, H.-T.; and Liu, T.-L. 2020.
\newblock Learning Gaussian Instance Segmentation in Point Clouds.
\newblock \emph{arXiv preprint arXiv:2007.09860}.

\bibitem[{Milioto et~al.(2020)Milioto, Behley, McCool, and
  Stachniss}]{milioto2020lidar}
Milioto, A.; Behley, J.; McCool, C.; and Stachniss, C. 2020.
\newblock LiDAR Panoptic Segmentation for Autonomous Driving.
\newblock In \emph{IROS}.

\bibitem[{Milioto et~al.(2019)Milioto, Vizzo, Behley, and
  Stachniss}]{milioto2019rangenet++}
Milioto, A.; Vizzo, I.; Behley, J.; and Stachniss, C. 2019.
\newblock RangeNet++: Fast and accurate LiDAR semantic segmentation.
\newblock In \emph{2019 IEEE/RSJ International Conference on Intelligent Robots
  and Systems (IROS)}, 4213--4220. IEEE.

\bibitem[{Pan et~al.(2021)Pan, Xia, Song, Li, and Huang}]{pan20213d}
Pan, X.; Xia, Z.; Song, S.; Li, L.~E.; and Huang, G. 2021.
\newblock 3d object detection with pointformer.
\newblock In \emph{IEEE Conf. Comput. Vis. Pattern Recog.}, 7463--7472.

\bibitem[{Paszke et~al.(2017)Paszke, Gross, Chintala, Chanan, Yang, DeVito,
  Lin, Desmaison, Antiga, and Lerer}]{paszke2017automatic}
Paszke, A.; Gross, S.; Chintala, S.; Chanan, G.; Yang, E.; DeVito, Z.; Lin, Z.;
  Desmaison, A.; Antiga, L.; and Lerer, A. 2017.
\newblock Automatic differentiation in PyTorch.
\newblock In \emph{NIPS-W}.

\bibitem[{Qi et~al.(2019)Qi, Litany, He, and Guibas}]{qi2019deep}
Qi, C.~R.; Litany, O.; He, K.; and Guibas, L.~J. 2019.
\newblock Deep hough voting for 3d object detection in point clouds.
\newblock In \emph{Int. Conf. Comput. Vis.}, 9277--9286.

\bibitem[{Qi et~al.(2017{\natexlab{a}})Qi, Su, Mo, and Guibas}]{qi2017pointnet}
Qi, C.~R.; Su, H.; Mo, K.; and Guibas, L.~J. 2017{\natexlab{a}}.
\newblock Pointnet: Deep learning on point sets for 3d classification and
  segmentation.
\newblock In \emph{IEEE Conf. Comput. Vis. Pattern Recog.}, 652--660.

\bibitem[{Qi et~al.(2017{\natexlab{b}})Qi, Yi, Su, and
  Guibas}]{qi2017pointnet++}
Qi, C.~R.; Yi, L.; Su, H.; and Guibas, L.~J. 2017{\natexlab{b}}.
\newblock Pointnet++: Deep hierarchical feature learning on point sets in a
  metric space.
\newblock In \emph{Adv. Neural Inform. Process. Syst.}, 5099--5108.

\bibitem[{Razani et~al.(2021)Razani, Cheng, Li, Taghavi, Ren, and
  Bingbing}]{razani2021gp}
Razani, R.; Cheng, R.; Li, E.; Taghavi, E.; Ren, Y.; and Bingbing, L. 2021.
\newblock GP-S3Net: Graph-based Panoptic Sparse Semantic Segmentation Network.
\newblock \emph{arXiv preprint arXiv:2108.08401}.

\bibitem[{Sirohi et~al.(2021)Sirohi, Mohan, B{\"u}scher, Burgard, and
  Valada}]{sirohi2021efficientlps}
Sirohi, K.; Mohan, R.; B{\"u}scher, D.; Burgard, W.; and Valada, A. 2021.
\newblock EfficientLPS: Efficient LiDAR Panoptic Segmentation.
\newblock \emph{arXiv preprint arXiv:2102.08009}.

\bibitem[{Smith(2017)}]{smith2017cyclical}
Smith, L.~N. 2017.
\newblock Cyclical learning rates for training neural networks.
\newblock In \emph{2017 IEEE winter conference on applications of computer
  vision (WACV)}, 464--472. IEEE.

\bibitem[{Tang et~al.(2020)Tang, Liu, Zhao, Lin, Lin, Wang, and
  Han}]{tang2020searching}
Tang, H.; Liu, Z.; Zhao, S.; Lin, Y.; Lin, J.; Wang, H.; and Han, S. 2020.
\newblock Searching Efficient 3D Architectures with Sparse Point-Voxel
  Convolution.
\newblock In \emph{Eur. Conf. Comput. Vis.}

\bibitem[{Thomas et~al.(2019)Thomas, Qi, Deschaud, Marcotegui, Goulette, and
  Guibas}]{thomas2019kpconv}
Thomas, H.; Qi, C.~R.; Deschaud, J.-E.; Marcotegui, B.; Goulette, F.; and
  Guibas, L.~J. 2019.
\newblock Kpconv: Flexible and deformable convolution for point clouds.
\newblock In \emph{Int. Conf. Comput. Vis.}, 6411--6420.

\bibitem[{Vaswani et~al.(2017)Vaswani, Shazeer, Parmar, Uszkoreit, Jones,
  Gomez, Kaiser, and Polosukhin}]{vaswani2017attention}
Vaswani, A.; Shazeer, N.; Parmar, N.; Uszkoreit, J.; Jones, L.; Gomez, A.~N.;
  Kaiser, {\L}.; and Polosukhin, I. 2017.
\newblock Attention is all you need.
\newblock In \emph{Adv. Neural Inform. Process. Syst.}, 5998--6008.

\bibitem[{Wang et~al.(2018)Wang, Girshick, Gupta, and He}]{wang2018non}
Wang, X.; Girshick, R.; Gupta, A.; and He, K. 2018.
\newblock Non-local neural networks.
\newblock In \emph{IEEE Conf. Comput. Vis. Pattern Recog.}, 7794--7803.

\bibitem[{Wang et~al.(2019)Wang, Liu, Shen, Shen, and
  Jia}]{wang2019associatively}
Wang, X.; Liu, S.; Shen, X.; Shen, C.; and Jia, J. 2019.
\newblock Associatively segmenting instances and semantics in point clouds.
\newblock In \emph{IEEE Conf. Comput. Vis. Pattern Recog.}, 4096--4105.

\bibitem[{Woo et~al.(2018)Woo, Park, Lee, and Kweon}]{woo2018cbam}
Woo, S.; Park, J.; Lee, J.-Y.; and Kweon, I.~S. 2018.
\newblock Cbam: Convolutional block attention module.
\newblock In \emph{Eur. Conf. Comput. Vis.}, 3--19.

\bibitem[{Xu et~al.(2020)Xu, Wu, Wang, Zhan, Vajda, Keutzer, and
  Tomizuka}]{xu2020squeezesegv3}
Xu, C.; Wu, B.; Wang, Z.; Zhan, W.; Vajda, P.; Keutzer, K.; and Tomizuka, M.
  2020.
\newblock Squeezesegv3: Spatially-adaptive convolution for efficient
  point-cloud segmentation.
\newblock In \emph{Eur. Conf. Comput. Vis.}, 1--19. Springer.

\bibitem[{Xu et~al.(2021)Xu, Zhang, Dou, Zhu, Sun, and Pu}]{xu2021rpvnet}
Xu, J.; Zhang, R.; Dou, J.; Zhu, Y.; Sun, J.; and Pu, S. 2021.
\newblock RPVNet: A Deep and Efficient Range-Point-Voxel Fusion Network for
  LiDAR Point Cloud Segmentation.
\newblock \emph{arXiv preprint arXiv:2103.12978}.

\bibitem[{Yang et~al.(2019)Yang, Wang, Clark, Hu, Wang, Markham, and
  Trigoni}]{yang2019learning}
Yang, B.; Wang, J.; Clark, R.; Hu, Q.; Wang, S.; Markham, A.; and Trigoni, N.
  2019.
\newblock Learning object bounding boxes for 3d instance segmentation on point
  clouds.
\newblock In \emph{Adv. Neural Inform. Process. Syst.}, 6740--6749.

\bibitem[{Ye et~al.(2021)Ye, Xu, Cao, and Chen}]{ye2021drinet}
Ye, M.; Xu, S.; Cao, T.; and Chen, Q. 2021.
\newblock DRINet: A Dual-Representation Iterative Learning Network for Point
  Cloud Segmentation.
\newblock arXiv:2108.04023.

\bibitem[{Zhang et~al.(2020)Zhang, Zhou, David, Yue, Xi, Gong, and
  Foroosh}]{zhang2020polarnet}
Zhang, Y.; Zhou, Z.; David, P.; Yue, X.; Xi, Z.; Gong, B.; and Foroosh, H.
  2020.
\newblock PolarNet: An Improved Grid Representation for Online LiDAR Point
  Clouds Semantic Segmentation.
\newblock In \emph{IEEE Conf. Comput. Vis. Pattern Recog.}, 9601--9610.

\bibitem[{Zhao et~al.(2020)Zhao, Jiang, Jia, Torr, and Koltun}]{zhao2020point}
Zhao, H.; Jiang, L.; Jia, J.; Torr, P.; and Koltun, V. 2020.
\newblock Point transformer.
\newblock \emph{arXiv preprint arXiv:2012.09164}.

\bibitem[{Zhou and Tuzel(2018)}]{zhou2018voxelnet}
Zhou, Y.; and Tuzel, O. 2018.
\newblock Voxelnet: End-to-end learning for point cloud based 3d object
  detection.
\newblock In \emph{IEEE Conf. Comput. Vis. Pattern Recog.}, 4490--4499.

\bibitem[{Zhou, Zhang, and Foroosh(2021)}]{zhou2021panoptic}
Zhou, Z.; Zhang, Y.; and Foroosh, H. 2021.
\newblock Panoptic-PolarNet: Proposal-free LiDAR Point Cloud Panoptic
  Segmentation.
\newblock In \emph{IEEE Conf. Comput. Vis. Pattern Recog.}, 13194--13203.

\bibitem[{Zhu et~al.(2020{\natexlab{a}})Zhu, Su, Lu, Li, Wang, and
  Dai}]{zhu2020deformable}
Zhu, X.; Su, W.; Lu, L.; Li, B.; Wang, X.; and Dai, J. 2020{\natexlab{a}}.
\newblock Deformable detr: Deformable transformers for end-to-end object
  detection.
\newblock \emph{arXiv preprint arXiv:2010.04159}.

\bibitem[{Zhu et~al.(2020{\natexlab{b}})Zhu, Zhou, Wang, Hong, Ma, Li, Li, and
  Lin}]{zhu2020cylindrical}
Zhu, X.; Zhou, H.; Wang, T.; Hong, F.; Ma, Y.; Li, W.; Li, H.; and Lin, D.
  2020{\natexlab{b}}.
\newblock Cylindrical and Asymmetrical 3D Convolution Networks for LiDAR
  Segmentation.
\newblock \emph{arXiv preprint arXiv:2011.10033}.

\end{thebibliography}
}

\end{document}